\documentclass[runningheads]{llncs}
\usepackage[T1]{fontenc}
\usepackage{array}
\usepackage{amssymb}
\usepackage{amsmath}
\usepackage{pifont}
\usepackage{bm}
\usepackage{xcolor}
\usepackage{booktabs} 
\usepackage{multirow} 
\usepackage[colorlinks=true,linkcolor=blue,citecolor=blue,urlcolor=blue,]{hyperref}
\usepackage{marvosym}
\usepackage{graphicx}
\begin{document}
\title{Reducing Annotation Burden: Exploiting Image Knowledge for Few-Shot Medical Video Object Segmentation via Spatiotemporal Consistency Relearning}
\titlerunning{Few-Shot Medical Video Object Segmentation with Images}
%
\author{Zixuan Zheng\inst{1}\thanks{Equal contribution.} \and
Yilei Shi\inst{1}$^\star$ \and
Chunlei Li\inst{1} \and
Jingliang Hu\inst{1} \and
Xiao Xiang Zhu\inst{2} \and
Lichao Mou\inst{1}\textsuperscript{(\Letter)}}


%
\authorrunning{Z. Zheng et al.}
%
\institute{MedAI Technology (Wuxi) Co. Ltd., Wuxi, China\\\email{lichao.mou@medimagingai.com} \and Technical University of Munich, Munich, Germany}
\maketitle              
\begin{abstract}
Few-shot video object segmentation aims to reduce annotation costs; however, existing methods still require abundant dense frame annotations for training, which are scarce in the medical domain. We investigate an extremely low-data regime that utilizes annotations from only a few video frames and leverages existing labeled images to minimize costly video annotations. Specifically, we propose a two-phase framework. First, we learn a few-shot segmentation model using labeled images. Subsequently, to improve performance without full supervision, we introduce a spatiotemporal consistency relearning approach on medical videos that enforces consistency between consecutive frames. Constraints are also enforced between the image model and relearning model at both feature and prediction levels. Experiments demonstrate the superiority of our approach over state-of-the-art few-shot segmentation methods. Our model bridges the gap between abundant annotated medical images and scarce, sparsely labeled medical videos to achieve strong video segmentation performance in this low data regime. Code is available at \url{https://github.com/MedAITech/RAB}.

\end{abstract}
\section{Introduction}
The emergence of deep learning has propelled the advancement of numerous medical data analysis tasks, particularly achieving significant accomplishments in segmentation tasks. The automatic segmentation of medical data plays a pivotal role in various clinical applications such as computer-aided diagnosis and disease progression monitoring~\cite{checvpr2023,che2023}. However, in many practical scenarios, segmentation models often face challenges due to the prohibitive cost of pixel-level annotation from clinical experts and the limited availability of samples for rare anomalies and unusual pathological conditions. To circumvent these constraints, there has been a growing interest in few-shot segmentation techniques that can learn to segment new classes with only a few annotated examples~\cite{suncbm2022}.
\par
Prior few-shot semantic segmentation methods in computer vision can be categorized into three main approaches: conditional networks, latent space optimization, and prototypical learning. Conditional networks commonly comprise two modules---a conditioning branch that ingests support samples to produce model parameters, and a segmentation branch that predicts masks for query images based on these dynamic parameters~\cite{rake2018,Shaban2017Arxiv}. Latent space optimization methods instead leverage generative models to synthesize images by optimizing latent vectors to match the distribution of queries. Segmentation is then performed by transferring representations from the synthesized examples~\cite{tritrong2021repurposing}. Finally, prototypical networks derive class prototypes from support images which are then compared against features of query images to predict segmentation masks~\cite{dongbmvc2018,PATNet}. In the medical domain, prototypical networks have garnered substantial traction for few-shot medical image segmentation~\cite{roymia2020,limia2023}.
\par
From image to video, \cite{chencvpr2021} and \cite{yancvpr2023} explore segmenting objects of novel categories in query videos using only a few annotated support frames. However, these approaches still require densely annotated frames for training and do not fully alleviate annotation costs. In addition, since videos are often not recorded and stored in medical imaging applications, publicly available medical video datasets remain relatively scarce and limited in scale, posing challenges for medical video object segmentation tasks.
\par
In this paper, we introduce a novel task of few-shot medical video object segmentation with image datasets. The goal is to segment a medical video given sparse annotations---only a few frames (e.g. the first) have ground truth masks, along with abundant labeled images. This represents an extremely low-data regime for video segmentation that minimizes costly pixel-level video annotations. To address this, we propose a two-phase framework. First, we learn a few-shot segmentation model using labeled images. Specifically, a pre-trained backbone extracts features from support and query images. These features and support masks are integrated to generate coarse query masks that approximately localize target objects. The coarse masks are then fused with support and query features to output fine segmentation masks. However, without temporal modeling, the few-shot image segmentation model may perform sub-optimally on videos. Thus, we propose a spatiotemporal consistency relearning approach for medical videos that is performed in the second phase. We leverage the temporal continuity prior, which assumes consistency between consecutive frames in a video, to obtain effective spatiotemporal features. Furthermore, we introduce spatial consistency constraints between the model trained in the first phase and the relearning model at both feature and prediction levels. These effectively minimize discrepancies between the two models to ensure model performance on videos. Our key contributions are three-fold:
\begin{itemize}
    \item We explore few-shot medical video object segmentation using image datasets, without reliance on full video annotations.    
    \item We introduce a novel self-supervised framework that exploits spatiotemporal consistency as additional supervision to achieve improved segmentations in this challenging setting.
    \item Extensive experiments demonstrate the superiority of our approach over existing state-of-the-art few-shot segmentation methods.
\end{itemize}

\begin{figure}
\centering
\includegraphics[width=\textwidth]{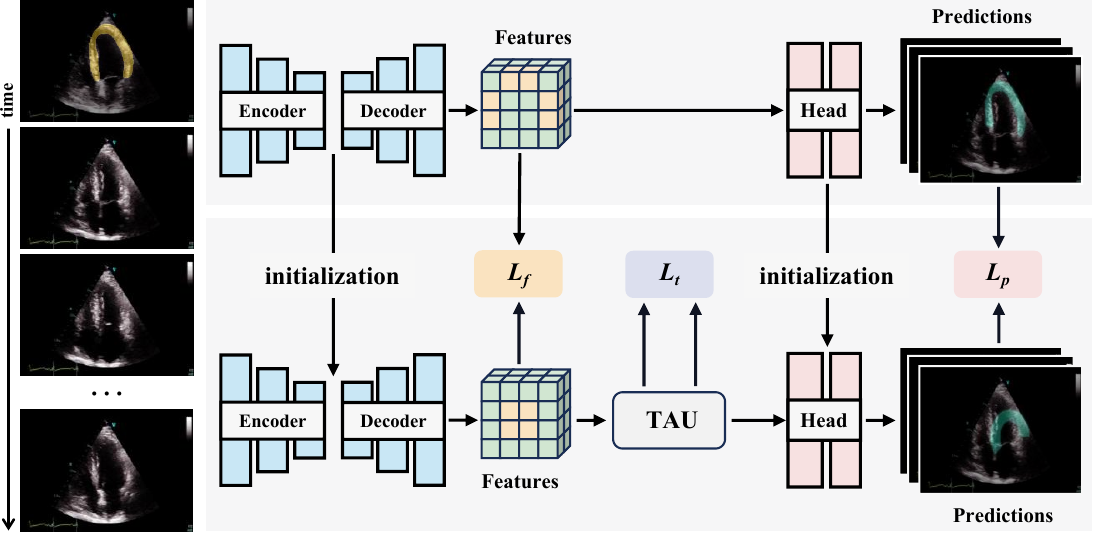}
\caption{Pipeline of the proposed few-shot segmentation method for medical video object segmentation. }
\label{fig_network}
\end{figure}

\section{Method}
\subsection{Task Definition}
We address the task of few-shot video object segmentation, where given annotated frames from the beginning of a medical video sequence, the goal is to segment subsequent frames. Our approach leverages labeled images to train a model, which is then applied to test videos containing unseen object classes. Both training and testing are performed in an episodic manner, where each episode contains data sampled from the same class. Specifically, an episode comprises a support set $\mathcal{S}=\{(\bm{s}_i,\bm{m}_i)\}_{i=1}^{N}$ and a query set $\mathcal{Q}=\{(\bm{q}_k,\bm{y}_k)\}_{k=1}^K$, where $\bm{s}_i$ and $\bm{m}_i$ represent the $i$-th support image and its ground truth mask, respectively, while $\bm{q}_k$ and $\bm{y}_k$ denote the $k$-th query image and the corresponding annotation mask. The model segments $\bm{q}_k$ guided by $\mathcal{S}$. For video data, we use the first $N$ frames as the support set and the remaining frames as query data.

\subsection{Learning Few-Shot Segmentation with Images}
We first train a few-shot medical image segmentation network that consists of three key components: a pseudo mask generation module, a cross-resolution feature fusion module, and a segmentation head. We extract visual features from support and query images using a pre-trained backbone. The pseudo mask generation module is designed to roughly localize query objects with minor cost. The cross-resolution feature fusion module then facilitates interaction between query features and support features along with pseudo masks at multiple scales, thereby preserving fine image details and precise localization cues. The output features from this module are fed into the segmentation head, comprising a $3\times3$ convolutional layer followed by a $1\times1$ convolution with softmax, to predict segmentation masks.

\subsubsection{Pseudo Mask Generation}
Pseudo masks have been widely adopted in few-shot segmentation models~\cite{Liu2022CVPR,Zhang2021NIPS} to provide coarse object localization. Specifically, they measure the cosine similarity between query and support features. We utilize high-level features to acquire pseudo masks, as they encode richer semantic information. Let $\mathcal{F}$ denote the backbone and $\bm{p}_k$ represent the generated pseudo mask for $\bm{q}_k$. We have
\begin{equation}
\label{eq:pseudo_mask}
\bm{p}_k=\cos\left(\mathcal{F}(\bm{q}_k),  \mathcal{F}(\bm{s}_i\odot\bm{m}_i)\right) \,,
\end{equation}
where $\odot$ is the Hadamard product. We then normalize $\bm{p}_k$ to the range of $\left[0,1\right]$ using min-max normalization.

\subsubsection{Cross-Resolution Feature Fusion}
First, we combine the resulting pseudo mask $\bm{p}_k$, containing localization cues, with support and query features as follows:
\begin{equation}
\bm{f}=\mathrm{Conv}_{1\times1}\circ\mathrm{Concat}(\mathrm{up}(\bm{p}_k),\mathcal{G}(\bm{q}_k),\mathcal{G}(\bm{s}_i\odot\bm{m}_i)) \,,
\end{equation}
where $\circ$ is a composition function, $\mathrm{up}(\cdot)$ upsamples the pseudo mask, $\mathcal{G}$ denotes the first half of the backbone network, and $\mathrm{Concat}(\cdot)$ represents channel-wise concatenation. Note that we utilize intermediate features (via $\mathcal{G}$) instead of high-level ones as in Eq.~(\ref{eq:pseudo_mask}), since they retain finer-grained details. Then, to derive multi-scale representations, we exploit the heavy neck module from GiraffeDet~\cite{Jiang2022ICLR}, and its output features are finally fed into the segmentation head to predict segmentation masks.

\subsection{Inference on Individual Video Frames  }
A straightforward approach to applying the above-mentioned few-shot segmentation model trained on images to videos is to treat each video frame as an independent sample, using the first few annotated frames as support data, with the remaining frames as individual query inputs.

However, empirical evidence suggests that this direct application does not yield optimal segmentation results. The fundamental issue arises from treating query data, i.e., video frames, as isolated inputs, inherently disregarding the temporal continuity inherent in video content.

\subsection{Spatiotemporal Consistency Relearning for Videos}

\subsubsection{Temporal Consistency} 
As mentioned above, the few-shot segmentation model, which is merely trained on images, fails to consider inter-frame correlations. A common approach to mitigate this issue, while reducing computational complexity, involves integrating new temporal modules with a frozen base model, such as adapters~\cite{adapter}. However, such methods typically require extra dense video annotations for parameter learning, which undermines the intention of few-shot learning in medical videos.
\par
To overcome these issues, we first introduce a temporal attention unit~\cite{TanCVPR2022} placed between the cross-resolution feature fusion module and segmentation head of the trained image model, designed to capture connections between frames. Then, we relearn this new model in a self-supervised manner. Under the temporal continuity prior, we utilize cosine similarity to regularize the consistency of features of consecutive frames. This serves to capture temporal relationships without supervision while improving segmentation.
\par
Formally, let $\bm{f}_t$ denote feature maps of the $t$-th frame from the inserted temporal attention module in a batch $\mathcal{B}$ which contains consecutive frames. We can have the following temporal consistency regularization:
\begin{equation}
\mathcal{L}_t=1-\frac{1}{\mathcal{|B|}} \sum_{t=1}^{|\mathcal{B}-1|}\cos(\bm{f}_t,\bm{f}_{t+1}) \,.
\end{equation}

\subsubsection{Feature and Prediction Consistency Constraints}
In the relearning phase, altering model parameters could cause the model to diverge from its training, thus deteriorating its performance on video data. For stability, we introduce the following two consistency constraints.
\par
The feature consistency implies that given the same inputs, features generated by the relearning model should remain correlated to the feature distribution of the trained, frozen image model. Let $\bm{z}_{i,j}$ and $\hat{\bm{z}}_{i,j}$ denote features from cross-resolution feature fusion modules of the two models, respectively. The feature consistency constraint can be defined as:
\begin{equation}
\mathcal{L}_f=\frac{1}{|\mathcal{B}|D} \sum_{t=1}^{|\mathcal{B}|}\sum_{j=1}^{N} (\hat{\bm{z}}_{i,j}-\bm{z}_{i,j})^2 \,,
\end{equation}
where $j$ is a spatial index, and $N$ denotes the number of feature vectors.
\par
The prediction consistency minimizes the discrepancy between segmentation masks predicted by the two models for the same frames. It can be formulated as:
\begin{equation}
\mathcal{L}_p=\frac{1}{|\mathcal{B}|} \sum_{t=1}^{|\mathcal{B}|}(\hat{\bm{y}}_t-\bm{y}_t)^2 \,.
\end{equation} 

\subsubsection{Overall Objective}
The weights of our relearning model, excluding the temporal attention unit, are initialized by the model pre-trained on images in the first phase. We fix the segmentation head and make the other parameters trainable. The overall objective for relearning is
\begin{equation}
\mathcal{L}=\lambda_1{\mathcal{L}_t}+\lambda_2{\mathcal{L}_f}+\lambda_3{\mathcal{L}_p} \,,
\end{equation}
where $\lambda_1$, $\lambda_2$, and $\lambda_3$ are trade-off hyperparameters. 
\par
Fig.~\ref{fig_network} shows our method. The shift from medical images to videos leads to performance degradation when deploying an image model on videos. The proposed relearning method enables the model to adapt to test video characteristics, mitigating this domain shift. By fine-tuning a subset of parameters, coupled with newly introduced ones, on a few labeled frames from the test data, the model can better capture the underlying distribution of the test data, improving performance.

\section{Experiment}
\subsection{Experimental Setups}
\subsubsection{Datasets}
For medical images, we consider the Breast Ultrasound Images (BUSI) dataset \cite{BUSI}, the Multi-Modality Ovarian Tumor Ultrasound (MMOTU) dataset \cite{MMOTU}, the TN3K dataset \cite{TN3K}, the Digital Database Thyroid Image (DDTI) dataset \cite{DDTI}, and the Laryngeal Endoscopic dataset \cite{vocaldata}.
For medical videos, we use the HMC-QU dataset \cite{heart} and the ASU-Mayo Clinic Colonoscopy Video (ASU-Mayo) dataset \cite{colonoscopy}.
To train a few-shot segmenter for ultrasound videos, we utilize BUSI, MMOTU, TN3K, and DDTI as training images, and perform relearning as well as testing on HMC-QU. Similarly, for experiments on endoscopic videos, we employ the Laryngeal Endoscopic dataset and ASU-Mayo. To enhance the performance of the image model learned in the first phase, we leverage a training strategy that combines natural image datasets, PASCAL-$5^i$~\cite{Shaban2017Arxiv} and COCO-$20^i$~\cite{Nguyen2019CVPR}, with the above-mentioned medical image datasets. Note that models are trained on base image classes and evaluated on novel video classes with no class overlap, assessing generalization to unseen data.

\subsubsection{Evaluation Metrics}
Following prior works~\cite{LiuECCV2020,Ouyang2020}, we adopt the Dice score and foreground-background IoU as evaluation metrics to assess the performance of few-shot segmentation models.

\subsubsection{Implementation Details}
For a fair comparison, all experiments are conducted under the most challenging one-shot setting, where only one annotated image (first phase) or frame (second phase) is used as the support set. We employ ResNet50~\cite{KM2016CVPR} pre-trained on ImageNet as our backbone network. During training of the few-shot medical image segmentation model, the backbone's weights are frozen except for block \#4, which is fine-tuned to learn more robust feature maps. The model is trained using the Adam optimizer with a learning rate of 1e-4 for the first 5K iterations, followed by the SGD optimizer with a learning rate of 1e-5 for subsequent iterations. The batch size is set to 8. In the second phase, we use the SGD optimizer with a learning rate of 1e-5 and a batch size of 4. Our method is implemented in PyTorch and runs on NVIDIA RTX 4090 GPUs.

\begin{table*}[!t]
\caption{Quantitative comparison with competing methods on the HMC-QU ultrasound video dataset for few-shot segmentation. Performance is reported in terms of Dice (\%) and IoU (\%).}
\label{tab:tableHMC-QU} 
\renewcommand\arraystretch{1.2}
\centering
\begin{tabular}{w{l}{2cm}| w{c}{1cm} w{c}{1cm}| w{c}{1cm} w{c}{1cm}| w{c}{1cm} w{c}{1cm}| w{c}{1cm} w{c}{1cm}}
\hline
\textbf{Images} & \multicolumn{2}{w{c}{2cm}|}{\textbf{Thyroid}} & \multicolumn{2}{w{c}{2cm}|}{\textbf{Breast}} & \multicolumn{2}{w{c}{2cm}|}{\textbf{Ovary}} & \multicolumn{2}{w{c}{2cm}}{\textbf{All}} \\
 \hline
 \hline
    methods & Dice & IoU & Dice & IoU & Dice & IoU & Dice & IoU \\
 \hline
    PATNet \cite{PATNet} & 32.91 & 21.82 & 32.57 & 20.37 & 35.65 & 22.44 & 36.61 & 22.76 \\
    HSNet \cite{HSNet} & 67.97 & 52.24 & 70.60 & 55.28 & 68.63 & 53.01 & 69.71 & 54.15 \\
    DCAMA \cite{DCAMA} & 60.09 & 41.02 & 75.61 & 61.33 & 63.03 & 46.52 & 68.75 & 52.99 \\
    DCP \cite{DCP} & 51.34 & 35.26 & 47.47 & 31.63 & 45.43 & 30.10 & 43.23 & 27.74 \\
    AFA \cite{AFA} & 72.36 & 57.41 & 75.11 & 60.63 & 67.99 & 52.26 & 65.03 & 49.26 \\
     Ours & \textbf{78.15} & \textbf{64.17} & \textbf{85.79} & \textbf{75.54} & \textbf{84.55} & \textbf{73.26} & \textbf{71.90} & \textbf{56.42} \\ 
 \hline
\end{tabular}
\end{table*}

\begin{table}[!t]
\centering
\begin{minipage}{0.45\textwidth}
\centering
\caption{Comparison with other methods on the ASU-Mayo endoscopic video dataset.}
\label{tab:tableASU-Mayo}
\renewcommand\arraystretch{1.2}
\setlength{\tabcolsep}{4pt}
\begin{tabular}{l|cc}
\hline
\textbf{Images}&\multicolumn{2}{c}{\textbf{Laryngeal}}\\
\hline
\hline
methods & Dice & IoU \\
\hline
PATNet \cite{PATNet} & 44.39 & 33.54 \\
HSNet \cite{HSNet} & 59.21 & 46.96 \\
DCAMA \cite{DCAMA} & 56.67 & 45.60 \\
DCP \cite{DCP} & 49.74 & 37.49 \\
AFA \cite{AFA} & 60.38 & 47.42 \\
Ours & \textbf{63.97} & \textbf{48.40} \\
\hline
\end{tabular}
\end{minipage}%
\hspace{0.5cm}
\begin{minipage}{0.45\textwidth}
\centering
\caption{Ablation studies on the effectiveness of each proposed consistency regularization term.}
\label{tab:ablation}
\renewcommand\arraystretch{1.2}
\setlength{\tabcolsep}{2pt}
\centering
\begin{tabular}{lll|cc|cc}
\hline
&&& \multicolumn{2}{|c}{\textbf{HMC-QU}} & \multicolumn{2}{|c}{\textbf{ASU-Mayo}} \\
\hline
\hline
$\mathcal{L}_t$ & $\mathcal{L}_f$ & $\mathcal{L}_p$ & Dice & IoU & Dice & IoU \\
\hline
\textbf{\ding{55}} & \textbf{\ding{55}} & \textbf{\ding{55}} & 83.55 & 71.89 & 62.69 & 48.05 \\
\textbf{\ding{55}} & \textbf{\ding{51}} & \textbf{\ding{51}} & 82.10 & 70.17 & 60.36 & 46.43 \\
\textbf{\ding{51}} & \textbf{\ding{55}} & \textbf{\ding{51}} & 83.37 & 71.26 & 61.85 & 47.65 \\
\textbf{\ding{51}} & \textbf{\ding{51}} & \textbf{\ding{55}} & 1.25 & 1.07 & 0.02 & 0.01            \\
\textbf{\ding{51}} & \textbf{\ding{51}} & \textbf{\ding{51}} & \textbf{85.79} & \textbf{75.54} & \textbf{63.97} & \textbf{48.4} \\
\hline
\end{tabular}
\end{minipage}
\end{table}

\subsection{Results} 
\begin{figure}[t]
\centering
\includegraphics[width=\textwidth]{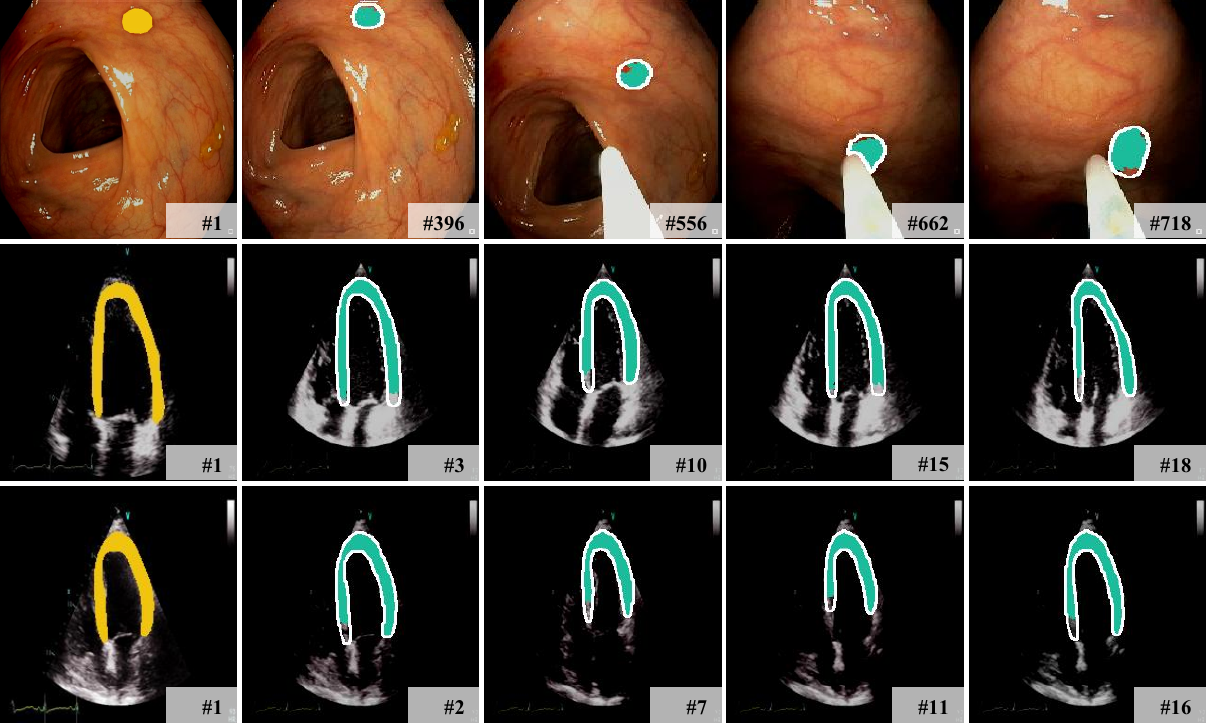}
\caption{Qualitative results of our few-shot video object segmentation model on the HMC-QU and ASU-Mayo datasets. The first column shows annotated support frames with ground truth masks (yellow). The remaining columns illustrate our model's segmentation predictions (green masks) on sampled query frames from videos. Ground truth masks for the query frames are outlined in white for reference.}
\label{fig_result}
\end{figure}

We compare the proposed model with state-of-the-art few-shot segmentation approaches, including PATNet~\cite{PATNet}, HSNet~\cite{HSNet}, DCAMA~\cite{DCAMA}, DCP~\cite{DCP}, and AFA~\cite{AFA}. PATNet and AFA utilize the prototype learning paradigm, whereas HSNet, DCAMA, and DCP introduce novel mechanisms to fully leverage information from support image-mask pairs. In Table~\ref{tab:tableHMC-QU}, ``thyroid'', ``breast'', and ``ovary'' denote the use of image datasets TN3K and DDTI, BUSI, and MMOTU, respectively, to train the few-shot medical image segmentation model. We also train the model with the combined ultrasound image datasets. In Table~\ref{tab:tableASU-Mayo}, ``laryngeal'' indicates the usage of the endoscopic dataset~\cite{vocaldata}. Experimental results in Table~\ref{tab:tableHMC-QU} and Table~\ref{tab:tableASU-Mayo} demonstrate that the proposed method outperforms competing approaches. The consistent superiority of our method across different datasets indicates its effectiveness and generalizability. Additionally,  qualitative results (see Fig.~\ref{fig_result}) show that our method generates better segmentation results for medical videos. Regarding time complexity, our model achieves 78 fps, comparable to baselines.

\subsection{Ablation Study}
We conduct ablation studies to investigate the effectiveness of each component in the proposed spatiotemporal consistency relearning approach. Table~\ref{tab:ablation} reports numerical results. We set the few-shot segmentation model pre-trained on images as the baseline. Then, we evaluate our relearning network under different loss combination settings. Compared to the baseline, lacking the temporal consistency regularization, i.e., $\mathcal{L}_t$, the Dice and IoU scores on both datasets decrease. Moreover, we observe that the constraints on feature consistency and prediction consistency, especially the latter, are essential for aligning distributions between the two phases, thereby guaranteeing model performance. By learning with all the introduced consistency losses, our model outperforms all competing approaches.

\section{Conclusion}
In conclusion, this work presents a novel approach to few-shot video object segmentation that bridges the gap between abundant annotated medical images and sparse video annotations. By first performing few-shot segmentation pre-training on labeled images and then introducing a spatiotemporal consistency relearning strategy on medical videos, our two-phase framework achieves strong performance in this extremely low-data regime without requiring costly dense video annotations. The relearning phase enforces consistency across consecutive frames while also maintaining constraints with the pre-trained image model. Experiments validate the superiority of our method over existing state-of-the-art few-shot segmentation techniques for video data. Overall, this approach provides an effective solution to leverage existing image labels for the video domain, minimizing costly new annotations. By enabling accurate video segmentation from just a few labeled frames, it opens up new possibilities for applications in the medical field where video data is prevalent but annotations are limited.

\begin{credits}

\subsubsection{\discintname}
The authors have no competing interests to declare that are relevant to the content of this paper.
\end{credits}
%
%
%
%

\end{document}